# AUTOMATIC FINANCIAL FEATURE CONSTRUCTION

*Jie Fang[1], Jianwu Lin[2], Yong Jiang[3], Shutao Xia[*]*

Tsinghua Shenzhen International Graduate School, Tsinghua University, Shenzhen, China
fangx18@mails.tsinghua.edu.cn, {lin.jianwu, jiangy, xiast}@sz.tsinghua.edu.cn

## ABSTRACT

In automatic financial feature construction task, the state-of-the-art technic leverages reverse polish expression to represent the features, then use genetic programming (GP) to conduct its evolution process. In this paper, we propose a new framework based on neural network, alpha discovery neural network (ADNN). In this work, we made several contributions. Firstly, in this task, we make full use of neural network's overwhelming advantage in feature extraction to construct highly informative features. Secondly, we use domain knowledge to design the object function, batch size, and sampling rules. Thirdly, we use pre-training to replace the GP's evolution process. According to neural network's universal approximation theorem, pre-training can conduct a more effective and explainable evolution process. Experiment shows that ADNN can remarkably produce more diversified and higher informative features than GP. Besides, ADNN can serve as a data augmentation algorithm. It further improves the the performance of financial features constructed by GP.

## 1. INTRODUCTION

In quantitative finance, useful features can help people find good trading opportunity. However, each feature has a capacity. If all people use the same feature, the opportunity will be quickly used up. Thus, more and more institutional investors start constructing new features from financial time series, and this topic has received more and more attention. Feature construction involves transforming a given set of input features to generate new and more powerful features, and these constructed features should be useful for prediction [1][2]. Most approaches rely on human experts to construct features, because it needs human to give an assessment on what is good and what is bad, and different industries have different criterions. What's more, it's easier to construct features from empirical data, other than raw data. However, constructing features by human experts is of low efficient. Thus, many researchers hope to find an automatic feature construction method. Sometimes feature construction and feature selection happens in the same procedure, so there are some methods can both do feature selection and feature construction. These methods are wrapper, filtering and embedded [3]. Filter is easy but of bad performance, it leverages specific criteria to choose features. Sometimes, it can help us to monitor the feature construction process. Wrapper performs well by directly using the model's results to serve as object function. So we can regard an individual trained model as a new constructed feature. However, it costs a lot of computation resource and time. Embedded is a method that uses generalized factors and a pruning technic to select or combine features, which serves as a middle choice between filtering and wrapper. In real practice, people prefer to use wrapper to achieve the desired performance.

The most frequently used wrapper method for automatic feature construction is genetic programming (*GP*) [4]. It uses tree structure to express the data and operators of explicit formulas. In each step, these trees will evolve, multiply and be selected, until they end up this process. Then the algorithm dynamically chose the right expression according to object function. Such as Lensen [5], Tran [6], Vafaie H [7], and Hindmarsh [8], uses *GP* to solve high dimension pixel data, and then produce new features for object detection task. They all uses the tree structure to represent a region of pixel, and then use *GP* to produce and selected trees. Finally, the last pixel tree is the produced feature.

There are also some researches which use neural network to represent new constructed features. Such as La [9], Liang [10] and Güvenir [11], uses convolutional neural network to represent pixel information in some area of a picture, then they use the trained classifier to serve as new feature. Virgolin [12] uses evolutionary meta learning approach in place of *GP* to construct features from medical data. Botsis T [13] leverages recurrent neural networks to build rule-based classifier among text data, each classifier represents a part of the text. They all have successfully used deep neural network to construct new features. With the development of deep neural network, we may have more opportunity to use wrapper approach. For some industries, such as health care, education and financial business, the empirical models are still much more powerful than only using deep learning algorithms [14][15][16]. Because these industries highly rely on people's behaviors and decisions, and deep neural network will easily over fit this nonstationary noise. Relatively, prior knowledge is more robust and reliable, because they really have tried to learn and understood this business. This point is what we highly valued in our work.

In quantitative investment area, people normally dig out financial features by their own thinking, economic knowledge or countless trials. They truly need automatic feature construction tools to increase their investment opportunity and cut down the costs. A very famous hedge fund company is the pioneer in this field. They successfully used genetic programming learner ($GP$) to automatically construct financial features [17]. Although they didn't officially publish papers to illustrate their algorithms, many people have tried to replicate this work. And similar approaches have been published. They use reverse polish expression to represent feature's formula, and use $GP$ to produce new expressions [18]. This method truly works, it have produced useful features to find right trading signals. However, the drawback is that it produced a lot of similar features and these features do not contain more information than the features produced by human experts.

In this paper, we regard this method as our baseline, and we want to find the pros and cons of our algorithms compared with this work [19]. We made several contributions compared with the works mentioned above. Firstly, we use deep neural network's wrapper structure to replace tree structure, and construct a reasonable object function to conduct this construction process. What's more, we come up with a kernel function to fix the objet function's undifferentiated problem. Secondly, we use model stealing technic to bring enough diversity into our network, which haven't tried by other researchers. Thirdly, we come out with a method to measure the diversity of different financial features. All these works make sure that we can automatically produce useful, highly informative and diversified features.

## 2. PRIOR KNOWLEDGE IN FINANCIAL SIGNAL PROCESSING

### 2.1. Problem Definition

In quantitative finance, the financial time series is our input data $X_{i,j}$ (i ∈ [1, n], j ∈ [1, T]). In $X_{i,j}$, i refers to $i_{th}$ types of data, such as open price, high price, low price, and volume, j refers different points of time. Take an example, now we are in the point t, we hope to construct a feature $f_t$ based on the data $X_{i,j}$ (i ∈ [1, n], j ∈ [1, t)). The feature $f_t$ contains a lot of stationary information in financial time series, and it should help us to forecast the near future. More specifically, in financial feature engineering, $f_t$ should has high spearman correlation with $R_t$. Here, $R_t$ refers to the return at point t. Its formula is shown as below:

$$R_t = \frac{\text{price}(t+\Delta t)}{\text{price}(t)} - 1 \quad (1)$$

In equity trading, institutional investors cared about the relative strength of each stock in each trading day. Thus, each batch should be made up by the data in the same trading day. Then, we use spearman correlation to calculate the correlation of feature value and future's return. This correlation can tell us, if the neural network proposes a feature value, how much money we can win or lose in the near future according to this value. The object function is shown in formula (2). w refers to the parameter in neural networks. What we want is to find the suitable w that can maximize this correlation.

$$\max_w C(f_t, R_t) = \frac{cov(rank(f_t - \overline{f_t}), rank(R_t - \overline{R_t}))}{var(rank(f_t - \overline{f_t})) * var(rank(R_t - \overline{R_t}))} \quad (2)$$

### 2.2. Network Structure

In our work, one of the biggest innovation points is to use deep neural network to do this task. Normally, neural network is hard to deal with the non-stationary data like financial time series. However, in this task, we want to find some stationary rules from the non-stationary data. Thus, neural network's strong feature extraction ability brings its overwhelming advantages. It can construct more informative features than other typical machine learning algorithms. This network structure is shown in figure 1.

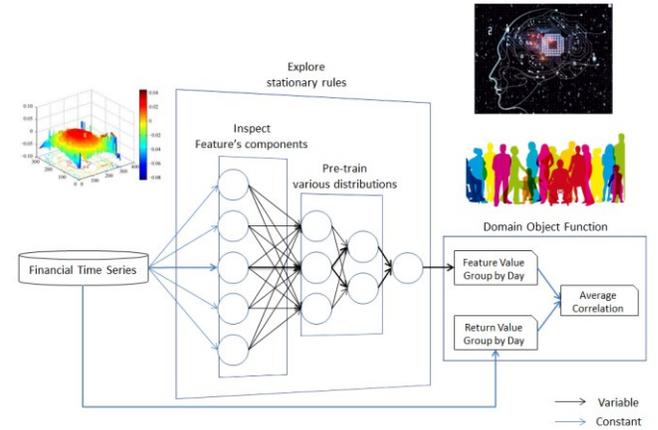

**Fig. 1.** Alpha discovery neural network's structure

Neural network requires all the operators are derivable. However, in formula 2, the operator rank() is not derivable. This operator is essential, because it can get rid of anomalies in data. Take an example, we have a list [-20,-2,-1,1,2,3,4,5,6,7,20]. Obviously, -20 and 20 are the anomalies in this data. The ideal situation is to project this series into [1,2,3,4,5,6,7,8,9,10,11]. Thus, we hope to construct a kernel function g(x) which can do this projection, and meanwhile this function should be derivable.

$$g(x) = \frac{1}{1+\exp(-p*\frac{x-\overline{x}}{2*std(x)})} \quad (3)$$

As shown in formula 3, at first, we project $x$ into a normal distribution which is zero-centralized. Next, we use hyper-parameter $p$ to make sure that the 2.5%-97.5% of data should lay in the range between $[mean - 2*std, mean + 2*std]$. In this way, we can get $p=1.83$. For example, one anomaly point $x_i = \overline{x} + 2*std$, solving $\frac{g(x_i)-g(\overline{x})}{g(\overline{x})} \leq \frac{x_i - \overline{x}}{\overline{x}}$, we can get std $\geq 0.362 * \overline{x}$. Which means, if one distribution's standard deviation is large, and it is larger than $0.362 * \overline{x}$, the g(x) can shorter the distance between anomalies and the central point. If the distribution's standard deviation is very small, g(x) will make it worse. However, even in this case, we can make sure that 95% of the points are between $[mean - 2*std, mean + 2*std]$, which is acceptable. And for financial time series, at the majority time, its distribution is not uniform.

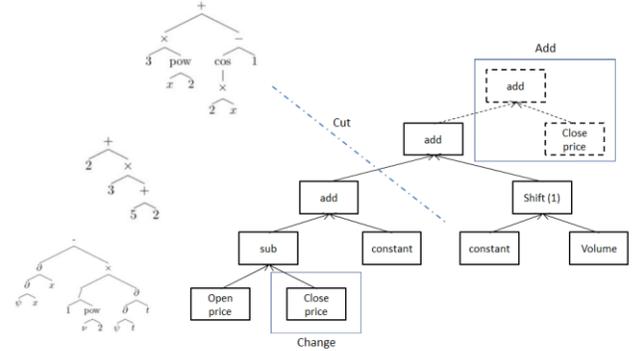

**Fig. 3.** This is the genetic programming's evolution process. It use reverse polish expression to represent financial features.

In GP, researchers add diversity into the constructed features by changing a part of the reverse polish expression. As a formula, a little bit change will make it totally different. Thus, this change may let the new feature totally have no relationship with the old feature. Although the old feature has high correlation with the return, no one can guarantee that the new feature still has high correlation with the return. The new feature can be a meaningless combination of operators and data. Thus, GP's evolution process is of low efficient, and it looks more like a searching process but not an evolution process.

In our method, each time, we pre-train our neural network with a classical financial feature. In this way, some part of the diversity can be permanently kept in the network [19]. After pre-training, we train this network with the object function mentioned above. According to neural network's universal approximation theorem [20], each time, the feature value changes slightly. In this way, the new feature always has close relationship with the old feature. With the help of right object function, the new feature will be always better than the old feature. Thus, we think the pre-train can provide us with a more efficient and intuitive evolution process.

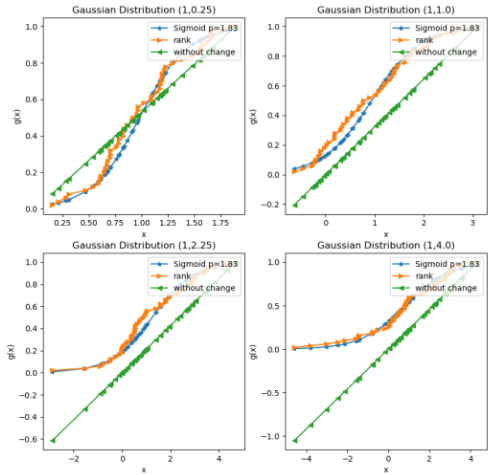

**Fig. 2.** We simulate 4 Gaussian distributions that have different mean and deviation. In different cases, we show how the kernel function $g(x)$ perform the same function like operator $rank()$

As we can see from figure 2, if the distribution's deviation is really small, there is no need to use g(x), but using g(x) will not make things worse. If the distribution's deviation is big, g(x) can almost perform the same function like rank(), and they have really big differences with the curve if we didn't do any changes.

By using deep neural network, and guiding its optimization direction with domain knowledge, we can easily get high informative financial features. However, we hope to find as much different features as possible. Thus, the second innovation point is that we use pre-train to replace the genetic programming. This evolution process is shown in figure 3.

Genetic programming can produce an explicit formula, which is highly explainable. But ADNN can only give us a feature value. We admit that it's ADNN's biggest shortcoming. For this shortcoming, we have consulted many professional institutional investors. Some of them do not care about it at all, because all these features are a small part of the system, they do not need to understand how it works, but they only need to make sure that it works well in the long run. Some of them think it's better to find a method to explain these features. At least, they should know the components of these features. Thus, we put forward a method, and try to make it explainable. We can fetch the weights between the input data and the first hidden state. We leverage $\sum_k |w_{j,k}|$ to represent the raw data j's contribution. In this way, the contribution of each raw data can be plotted, shown in figure 4.

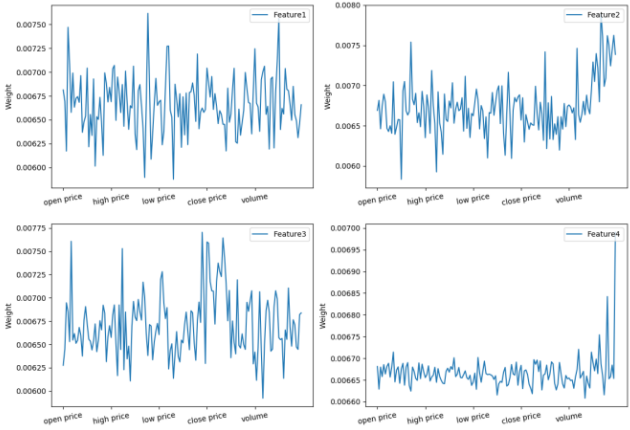

**Fig. 4.** Inspect different features' components by visualizing the raw data's contribution

By using this method, we can visualize one feature's evolution process, shown in figure 5.

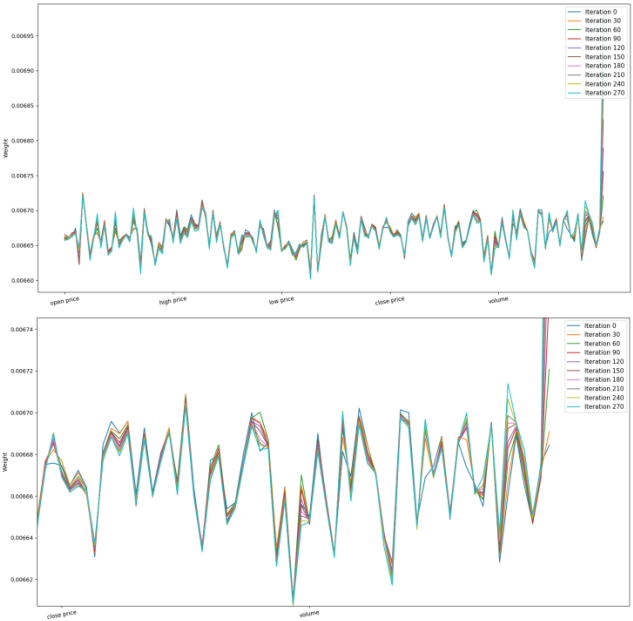

**Fig. 5.** Visualize ADNN's evolution process. Every step is continuous, thus it's more like a real evolution process but not a searching process

As we can see from figure 5, the raw data's contribution has changed slightly. That's why we say the initialization's differences can be permanently kept in in the network. Meanwhile, some parameters have been changed, which aims at improving the performance of input financial feature.

Besides the correlation, features' diversity is another very important indicator to measure the algorithm's performance. We propose several methods to quantify the produced features' diversity [21]. In step 1, we use our method to produce $m$ inexplicit formulas. If there are $n$ stocks, each formula will produce a $1*n$ array to represent feature distribution in a given trading day. As mentioned before, we only care about the features' relative strength in each trading day. Thus, we do *softmax* on feature's distribution in each trading days. This step can help us ignore data's scale without losing any information. In step 2, we measure the distance between two features' distribution. $m$ formulas can produced a $m*m$ distance matrix. In step 3, we use k-means to do clustering on these formulas. Clustering can help us get a more stable diversity, ignoring the anomalies. What's more, when the cluster center's number is approximately 5%-15% of the sample number, the results are similar. In step 4, we calculate the average distance between each cluster center. This value can represent our method's diversity in a given trading day.

In step2, there are several common methods to measure the distance of different financial features [21], which is Euclidean Distance, Cross Entropy, 1-cos(x) and 1-correlation. Here, we leverage different methods to measure the diversity of ADNN and GP.

**Table. 1.** We use different methods to measure the distance. The average diversity is calculated from testing set, 50 features, from May 2019 to July 2019.

| Methods | ADNN | GP |
|---|---|---|
| Euclidean Distance | 0.088 | 0.028 |
| Cross Entropy | 25.64 | 18.30 |
| 1-Cos(x) | 0.574 | 0.392 |
| 1-Correlation | 0.716 | 0.525 |

As we can see from Table 1, in all methods, ADNN' diversity is larger than GP. Thus, we think this result is robust. Due to the fact that cross entropy is more frequently used in neural networks to measure two distributions' distance, thus, we only use cross entropy to measure the distance in the following experiments.

## 3. EXPERIMENTS

### 3.1. Experiment Setting

In numerical experiment, we use A-share stock market's day frequency data, and construct daily technical factors (the financial features constructed by price and volume). When we construct the features and do backtest, we didn't take the pre-IPO, breakdown, and any other un-tradable stocks into account.

For each constructed feature, its input data is open price, high price, low price, close price and volume in the past 30 trading days. Because in each batch, there is 10*3700 samples. At present, we can only tolerate 30 trading days, and it needs 20g GPU's memory. Besides, most of the technic factors are made up by the nearest price and volume, which is the latest one week's data. Thus, this experiment setting is reasonable, but there is room for further improvements. And more computing resource or good parallel computing technic will be helpful for this task.

We calculate the mean value and standard deviation of each feature in training set. Then we use *(x-mean)/std* to standardize the data. In this way, we can get the input tensor as batch size *150. The return is calculated by the next 5 trading day, it is a batch size *1 tensor. For each batch, we randomly select 10 trading days. This sampling rule can make our network very robust, and find the stationary rules buried in the non-stationary financial time series. Then we calculate the spearman correlation between the feature value and the return in each trading day. At last, the mean value of this 10 day's spearman correlation serves as object function.

For each experiment, the first 250 trading days serve as training set, the following 30 trading days serve as validation set, and the following 30 trading days serve as testing set. Because no technical factors can work well all the time in the financial time series, it's unreasonable to enlarge our training set. And 250 trading days is quite a long period of time, in each trading day, there is more than 3000 samples. The reason we choose such a long period of time is that it can help us find the stable but not outstanding features in the beginning. Then, we use its performance in the validation set to conduct an early stopping. In this way, we can find the features which are the most active and powerful in the latest time, among the stable features. The robust gradient process is plotted in figure 6.

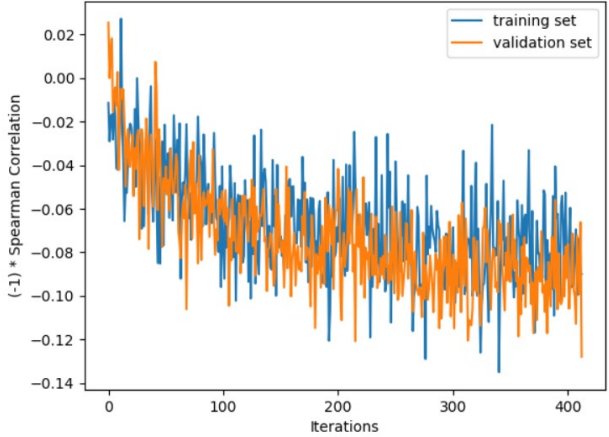

**Fig. 6.** The financial features perform differently in each trading day. As a consequence, the loss fluctuates greatly. The gradient process is clear, which means ADNN steadily explores informative features.

### 3.2. Outperform Genetic Programming

We use the same experiment setting for the genetic programming algorithm (*GP*). In order to quantify and assess our algorithms' performance, we propose three schemes as following:

**Table 2.** Three different schemes to show ADNN's contributions, compared with baseline.

| |
| --- |
| **Scheme A:** Only use *GP* to construct 100 features. |
| **Scheme B:** Use *GP* to construct 100 features at first, and then uses these features as prior knowledge to pre-train our network. Finally, we get 100 features by *ADNN*. |
| **Scheme C:** Only use *ADNN* to construct 100 features. Each time, we pre-train the network with randomly produced distribution. |

For each trading day, we compare the average spearman coefficient (it also called information coefficient, IC) and diversity among 100 produced features. In the same time, we use the same experiment settings to run *GP*. And get its results for Scheme B and Scheme C. The performance for each scheme is shown in figure 7.

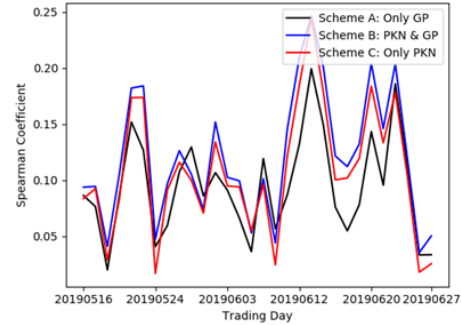

**Fig. 7.** Scheme A, Scheme B and Scheme C's performance on given trading days. All results are from testing set

Due to the non-stationary characteristics of financial data, spearman coefficient is not very stable in each trading day. However, as we can see from figure 4, we can find that Scheme B's spearman coefficient and diversity is relatively better than Scheme A. And Scheme C is also relatively better than Scheme A. The summary of the results is shown in Table 2, which is calculated from mean value of all trading days in testing set.

**Table 3.** Summarize different schemes' performance. We also list how much time (hour) to construct 50 features.

| Object | Test IC | Test Diversity | Time |
| --- | --- | --- | --- |
| Scheme A(Only *GP*) | 0.093 ± 0.035 | 17.16 ± 2.66 | **0.242** |
| Scheme B (*GP&ADNN*) | **0.121** ± 0.031 | **24.34 ± 5.68** | 1.155 |
| Scheme C(Only *ADNN*) | 0.110 ± 0.010 | 23.85 ± 4.17 | 0.837 |

All the results shown in Table 3 are from testing set. If a feature's test spearman coefficient is larger than 0.05, we normally regard it as a good trading indicator. For diversity, because we put forward a new method to measure it. We can't comment their performance by its value, we can only compare their relative performance.

Comparing each scheme's spearman coefficient, we can get Scheme C > Scheme A. It shows that *ADNN* can dig out more informative features than *GP*. What's more, Scheme B > Scheme C > Scheme A shows that the quality of pre-trained prior knowledge will also influence its final achievements. Thus, if we have good prior knowledge, with the help of *ADNN*, we can always produce good features, just like Scheme B.

Comparing each scheme's diversity, we find *GP*'s diversity is the worst, which is its well-known shortcoming. Although these features have different explicit formulas, they can be easily classified into the same category. *ADNN* can produce relatively more diversified features, and we think *model stealing* is the right method to add diversity. For Scheme B > Scheme C, because Scheme B is pre-trained with randomly produced distribution. Most of these distributions can't serve as useful financial features. Thus, *ADNN* can't carry on their initial direction which let it can't enjoy the diversity brought by different distributions. For the features produced by *GP*, although their diversity is relatively low, they are truly useful features in financial time series. Thus, if we pre-train with these features, these features' differences can positively be passed to our network.

### 3.3. More Experiments

Due to the fact that financial time series is extremely non-stationary [22], although our method works well from 2018 to 2019, it can't promise to work well in the other time. Thus, we repeated our experiments several times. Here is Scheme B's average performance in the last four years, shown in Table 4.

**Table 4.** Scheme B's performance in the last four years, value comes from testing set.

| Time | Test IC | Test Diversity |
|---|---|---|
| Train: Jan 2015 to Dec 2015<br>Test: Feb 2016 to May 2016 | 0.117 ± 0.028 | 20.93 ± 8.32 |
| Train: Jan 2016 to Dec 2016<br>Test: Feb 2017 to May 2017 | 0.132 ± 0.025 | 21.22 ± 7.26 |
| Train: Jan 2017 to Dec 2017<br>Test: Feb 2018 to May 2018 | 0.081 ± 0.045 | 29.05 ± 4.13 |
| Train: Jan 2018 to Dec 2018<br>Test: Feb 2019 to May 2019 | 0.121 ± 0.031 | 24.34 ± 5.68 |

As we can see from Table 4, the performance in Jan - Dec 2017 is not good. It's acceptable, because the majority features constructed by using price and volume in this year doesn't work well. But all the test spearman coefficients are higher than 0.05, and all the test diversity are higher than Scheme A shown in Table 2. Thus, we can regard *ADNN* as a reliable method which can be put into industry use.

For financial automatic feature construction task, it's enough to assess the algorithms' performance according to the information coefficient and diversity, in testing set. But if we want to put it into real use, we have to deal with the feature selection process, which is not the research topic in this work. However, in this work, we find that ADNN can construct more informative and more diversified features than GP. Thus, if we can still uses the three schemes in table 1 to compare the performance of ADNN and GP. Here, we only make a small change. For each algorithm, we only construct 5 features for each scheme, because most of quantitative strategies adopt less than 20 technical factors. For this quantitative strategy, we only invest in the tradable stocks in A-share market. Every week, we use XGBoost to make a binary classification on the stocks, based on 5 features constructed by each scheme. We long the top 10% stocks (A-share market doesn't support shorting stocks), and hold this portfolio for 5 trading days. The result is shown in Table 5.

**Table 5.** The value in the table is the strategy's absolute return, from the testing set. ZZ500 is very important index in A-share market.

| Time | Scheme A | Scheme B | Scheme C | ZZ500 |
|---|---|---|---|---|
| Train: 2015.01-2015.12<br>Test: 2016.02-2016.03 | +2.59% | +5.74% | +4.52% | +1.67% |
| Train: 2016.01-2016.12<br>Test: 2017.02-2017.03 | +5.40% | +10.26% | +8.33% | +2.53% |
| Train: 2017.01-2017.12<br>Test: 2018.02-2018.03 | -5.27% | -4.95% | -4.16% | -6.98% |
| Train: 2018.01-2018.12<br>Test: 2019.02-2019.03 | +13.00% | +15.62% | +15.41% | +13.75% |

Because this strategy is based on 5 technical factors, its performance is not outstanding. We want to compare these schemes' relative strength, to show the relative strength of ADNN and GP. As shown in table 5, in all periods of time, Scheme C > Scheme A, which means the features constructed by ADNN can find better trading opportunity than the features constructed by GP. At most of the time, Scheme B is the best, which means a good initialization is very important. A good initialization can make positive impact on the final achievement of a neural network. Moreover, with the help of ADNN, the performance of GP can be further improved.

### 4. CONCLUSION

Different from previous works, we put forward a framework to automatically construct financial features based on neural network. In this framework, we make full use of neural network's overwhelming advantage in feature extraction. Moreover, according to neural network's universal approximation theorem, we leverage pre-training to conduct a more effective and explainable evolution process, than genetic programming. The experiment result shows that *ADNN* can construct more informative and diversified features than the state-of-the-art technic *GP*. And ADNN can also serve as a data augmentation algorithm to improve the performance of existing technical indicators.